\documentclass{article}

\usepackage[attention,final]{neurips_2021}

\usepackage[utf8]{inputenc} 
\usepackage[T1]{fontenc}    
\usepackage{hyperref}       
\usepackage{url}            
\usepackage{booktabs}       
\usepackage{amsfonts}       
\usepackage{nicefrac}       
\usepackage{microtype}      
\usepackage{xcolor}         
\usepackage{graphicx}
\usepackage{amsmath}
\usepackage{subcaption}

\title{Neurosymbolic Programming for Science}

\author{%
  Jennifer J. Sun\thanks{Equal contribution} \\
  Caltech\\
   \And
   Megan Tjandrasuwita\footnote[1]{} \\
   MIT CSAIL \\
   \And
   Atharva Sehgal\footnote[1]{} \\
   UT Austin \\
   \AND
   Armando Solar-Lezama \\
   MIT CSAIL \\
   \And
   Swarat Chaudhuri \\
   UT Austin \\
   \And
   Yisong Yue \\
   Caltech \\
   \And
   Omar Costilla-Reyes \thanks{Corresponding author: costilla@mit.edu} \\
   MIT  CSAIL \\
}

\begin{document}

\maketitle

\begin{abstract}
Neurosymbolic Programming (NP) techniques have the potential to accelerate scientific discovery. These models combine neural and symbolic components to learn complex patterns and representations from data, using high-level concepts or known constraints. NP techniques can interface with symbolic domain knowledge from scientists, such as prior knowledge and experimental context, to produce interpretable outputs. We identify opportunities and challenges between current NP models and scientific workflows, with real-world examples from behavior analysis in science: to enable the use of NP broadly for workflows across the natural and social sciences.   
\end{abstract}

\section{Introduction}

One of the grand challenges in the artificial intelligence and scientific communities is to find an AI scientist: an artificial agent that can automatically design, test, and infer scientific hypotheses from data. This application poses several distinct challenges for existing learning techniques because of the need to ensure that new theories are consistent with prior scientific knowledge, as well as to enable scientists to reason about the implications of new hypotheses and experimental designs.

The distinct requirements of scientific discovery have pushed the community to explore expressive yet symbolically interpretable techniques such as symbolic regression  \citep{cranmer2020pysr}, interpretable machine learning  \citep{ustun2017optimized, doshi2017towards, kleinberg2018human, mcgrath2021acquisition}, as well as program synthesis  \citep{koksal2013synthesis,ellis2022synthesizing}. These techniques have helped the community make significant progress in a number of applications, such as those discussed in ~\cite{goodwin2022toward} and \cite{sapoval2022current}, but we are still far from solving the grand challenge. 

We focus on the opportunities and challenges behind an important class of learning techniques based on \emph{Neurosymbolic Programming} (NP)~\citep{chaudhuri2021neurosymbolic}. These techniques combine neural and symbolic reasoning to build expressive models that incorporate prior expert knowledge and strong constraints on model behavior and structure. NP is capable of producing symbolic representations of theories that can be analyzed and manipulated to answer rich counterfactuals. 

\begin{figure}
    \centering
    \includegraphics[width=\linewidth]{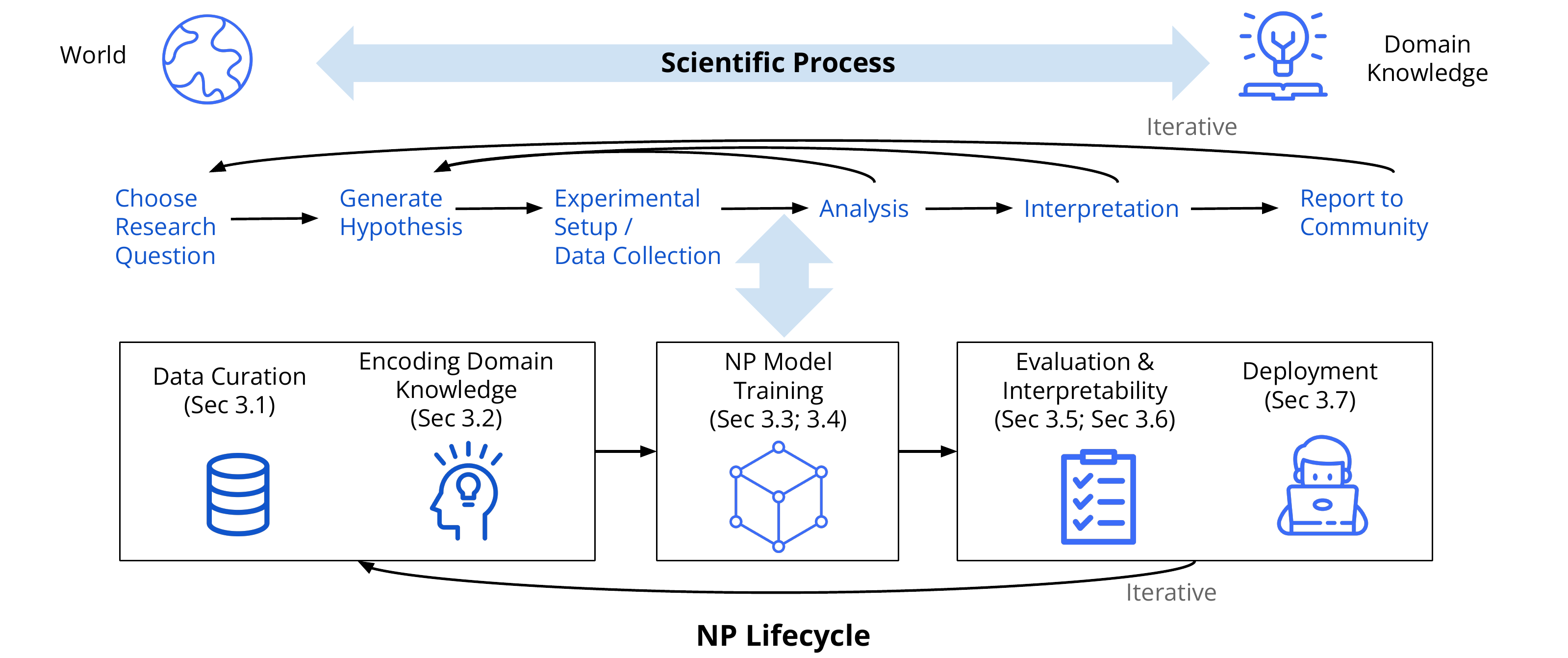}
    \vspace{-0.175in}
    \caption{Synergy between the scientific and neurosymbolic programming workflow.}
    \vspace{-0.175in}
    \label{fig:process_example}
\end{figure}

NP empowers a new line of attack on the grand AI scientist challenge: represent scientific hypotheses as programs in a \textit{Domain Specific Language} (DSL) and use neurosymbolic program synthesis to automatically discover these programs (Figure \ref{fig:process_example}). Users can incorporate complex prior knowledge (e.g., known features and constraints) into the design of the DSL. The NP learning algorithms can then follow classic scientific reasoning principles to find predictive programs. Also, models learned this way are often similar to code that human domain experts write during manual scientific modeling. Collectively, these characteristics enable a transparent and interactive process where an AI system and a human expert collaborate on evidence-based reasoning and the discovery of new scientific facts.

Here, we use behavior analysis as a concrete, illustrative example. We start with an introduction to NP (Section~\ref{sec:NP_techniques}), then outline challenges and opportunities for future research (Section~\ref{sec:opp}).

\textbf{Behavior analysis as running example.} We chose behavior analysis as an example use case for several reasons.
Behavioral data is spatiotemporal, which is a common data type across the sciences.
Correspondingly, underlying challenges are shared in other domains, from monitoring vital signs to modeling physical systems, to studying the dynamics of chemical reactions.
Additionally, behavioral data illustrate common challenges with scientific data. 
These datasets often contain rare behaviors with noisy and imperfect data and can vary significantly in relevant time scales (e.g., milliseconds vs hours). Datasets also vary across labs, organisms/systems, and experimental setups. 
Finally, automatic behavior quantification is becoming increasingly crucial in many fields, such as neuroscience, ecology, biology, and healthcare. 
As computational behavior analysis and neurosymbolic learning are both developing research areas, there are many exciting opportunities to explore at their intersection.
    
\textbf{Background on behavior analysis.} An important objective of behavior analysis is to quantify behavior from video using continuous or discrete representations. We focus on the case of animal behavior analysis in science~\citep{anderson2014toward,datta2019computational}, where there are diverse organisms and naturalistic behaviors.
A common approach is first to perform animal pose tracking from video~\citep{Mathisetal2018,Pereira2022sleap}, then categorize behaviors of interest from animal pose~\citep{segalin2021mouse} (as discussed later in Figure \ref{fig:mars_pipeline}).  From an NP perspective, this approach can be viewed as learning a symbolically interpretable intermediate representation (tracked keypoints).

\textbf{Existing challenges in behavior analysis.} Similar to other scientific fields, data collection and annotation are expensive for behavioral experiments.
Analyzing data is also time-consuming and expensive since specialized domain expertise is required for identifying behaviors of interest and extracting knowledge.
Models need to interface efficiently with scientists and data at both the inputs and outputs from the scientific process (Figure~\ref{fig:process_example}).
For NP models, leveraging domain expertise in the form of behavioral attributes has been demonstrated to improve data efficiency~\citep{sun_task_2021} and interpretability~\citep{tjandrasuwita2021interpreting}.

There is a variety of domain expertise that requires new algorithmic designs to integrate into the NP workflow, such as experimental context, existing ethograms, and scientific spatiotemporal constraints.

\begin{figure}
        \centering
        
\begin{subfigure}[b]{\textwidth}
\centering
   \includegraphics[width=0.75\linewidth]{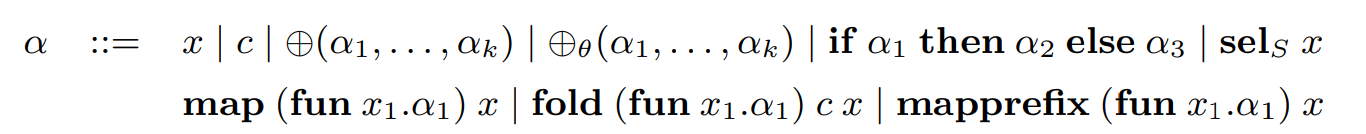}
   \caption{Example of a domain-specific language for neurosymbolic programming.}
   \label{fig:Np1} 
\end{subfigure}        
\begin{subfigure}[b]{\textwidth}

\centering
   \includegraphics[width=0.85\linewidth]{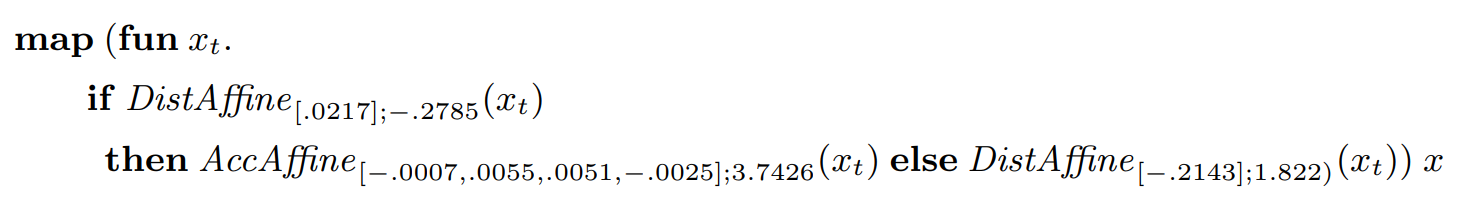}
   \caption{A neurosymbolic program.}
   \label{fig:Np2}
\end{subfigure}
        \vspace{-0.15in}
        \caption{Examples of NP for learning programs in mouse social behavior ~\citep{shah_learning_2020}.}
    \vspace{-0.175in}
        \label{fig:near_program}
    \end{figure}

Incorporating such domain knowledge has the potential to enable NP models to be more robust to noisy and imperfect data, and enable new scientific inquiries that were too expensive to study previously.
Furthermore, when black-box models are used for studying behavior, it is difficult to diagnose errors and explain model outputs~\citep{rudin2019stop}. 
NP models have the potential to produce symbolic descriptions of behavior (Figure~\ref{fig:near_program}), which enables experts to connect model interpretations with other parts of the behavior analysis workflow, e.g., describing behavioral differences across different strains of mice.
Finally, to enable the use of NP models in real-world science workflows, these models must be scalable and produce robustly reproducible interpretations. 
    
    \begin{figure}[b]
        \makebox[\textwidth][c]{\includegraphics[width=\textwidth, height=4.75cm]{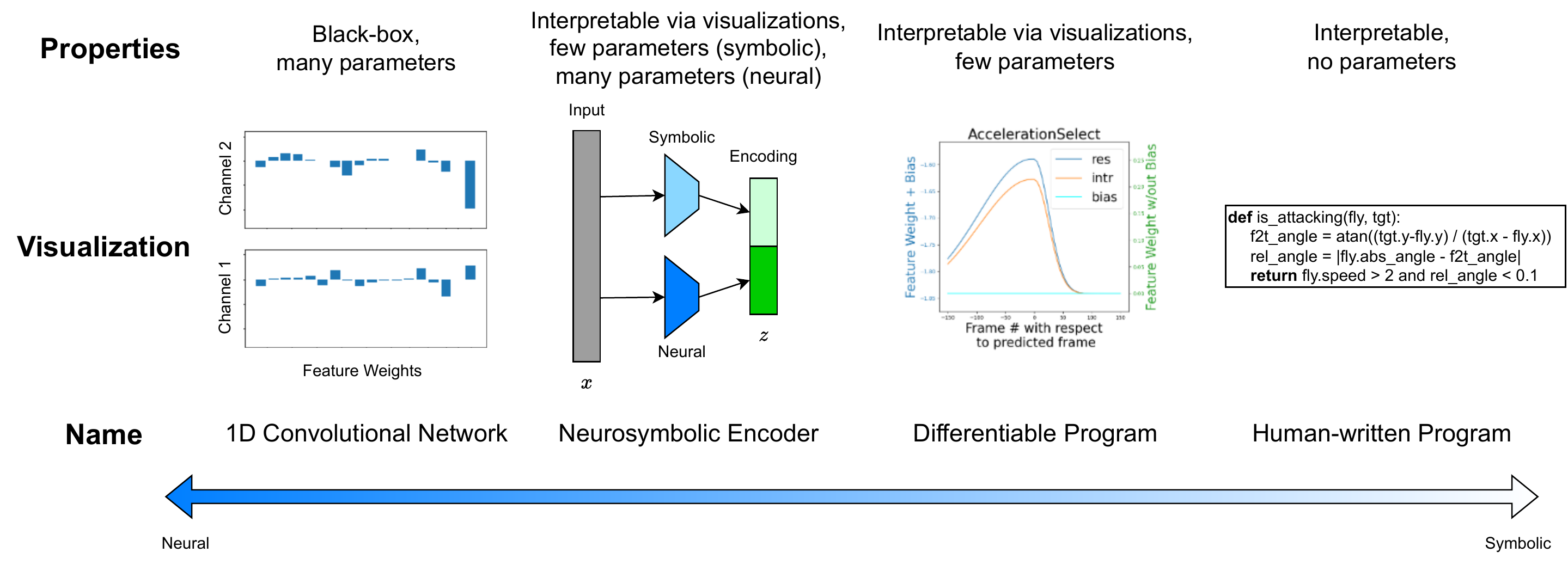}}
        \vspace{-0.25in}
        \caption{Space of neurosymbolic programming models in behavior analysis, including purely neural (left), purely symbolic (right), and neurosymbolic (two in middle)}
        \label{fig:1d_plot}
    \end{figure}

\section{Neurosymbolic Programming Techniques}\label{sec:NP_techniques}
\emph{Neurosymbolic programs} incorporate latent representations from neural networks and symbols that explicitly capture pre-existing human knowledge, and connect these elements using rich architectures that mirror classic models of computation. The programs, assumed to belong to a DSL, are learned using a combination of gradient-based optimization, probabilistic methods, and symbolic techniques. 

\textbf{Anatomy of a Neurosymbolic Program.}
In general, a neurosymbolic program comprises its discrete architecture and continuous parameters. Consider the neurosymbolic program in Figure \ref{fig:Np2}, obtained from DSL \ref{fig:Np1}, which comprises logical symbolic operations such as ``if'' statements, as well as functions with continuous parameters.  The architecture includes all the discrete symbolic choices that form the structure of the program (such as whether to have an ``if'' statement and where to place it relative to other operations), and ``programming'' this architecture is analogous to architecture design in neural networks (e.g., whether to use convolutions, recurrent units, attention, etc.).

\textbf{Space of Neurosymbolic Programs.}
The range of NP methods varies in the degree to which they use neural versus symbolic reasoning (Figure \ref{fig:1d_plot}).
The two ends of the spectrum correspond to purely neural (a 1D convolutional network) and purely symbolic (a human-written program) models, respectively. The techniques close to the center are neurosymbolic: the model in the center-left is a neurosymbolic encoder \citep{zhan2021unsupervised}, while the model in the center-right is a program with differentiable parameters for behavior analysis \citep{tjandrasuwita2021interpreting}.  From a definitional perspective, purely neural and purely symbolic programs can be considered special cases of neurosymbolic programs, although we typically do not refer to those as neurosymbolic programs for practical purposes.

To illustrate the strengths and weaknesses of each model in Figure \ref{fig:1d_plot}, assume that we have a scientific hypothesis to test on a dataset. On the right side, the fully symbolic model would involve an expert-written program that encodes the hypothesis in a general programming language. This program requires no learnable parameters, is fully interpretable and, if needed, can be iteratively improved. However, this method is also brittle, and the program must be engineered to handle all the dynamics of the dataset. This is intractable for models with complex dynamics. On the left side, the purely neural model would model the hypothesis directly using the dataset. Such models fit well to the dataset but offer limited interpretability and control over the generated hypothesis, which can make them prone to overfitting and limit generalization. In the middle, neurosymbolic approaches offer a mixture of both neural and symbolic components, and if designed well can inherit both the flexibility of neural networks as well as the structured semantics of symbolic models.  

\textbf{Desiderata for Effective Neurosymbolic Programming.} There are two main requirements for effective neurosymbolic programming: having a good DSL, and scalable learning algorithms.  The DSL effectively corresponds to the component building blocks from which one can construct a neurosymbolic architecture, and is one of the primary ways that experts inject domain knowledge (i.e., inductive bias) into the learning process.  Learning has two aspects: 1) searching for the best neurosymbolic architecture, and 2) optimizing the parameters within a fixed architecture.  The former is is analogous to neural architecture search \cite{elsken2019neural}, and the latter is analogous to standard parameter optimization in deep learning.  We discuss these issues in detail in Section \ref{sec:opp}, and conclude this section with a discussion on differentiable programs that enable differentiable parameter optimization within discrete symbolic architectures.

\textbf{Differentiable programs.} A differentiable program for a programming language is defined as a composition of functions such that the parameters of the function  are differentiable. A differentiable program follows the syntax defined by a DSL which can consist of parametric functions (multi-layered perceptrons, linear transformations), algebraic functions (\texttt{add}, \texttt{multiply}), and programming languages higher-order functions (\texttt{map}, \texttt{fold}). Furthermore, the composition of these differentiable functions is also differentiable through the chain rule. This property enables resulting programs to be fully differentiable. 
 
NP programs may be difficult to interpret by domain experts ~\citep{tjandrasuwita2021interpreting}  focuses on explaining the difference in behavior expert annotations. They replace generic higher-order functions over recursive data structures, e.g. \texttt{map} and \texttt{fold}, with a differentiable temporal filter operation, the Morlet Filter. The filter models temporal information in a highly data-efficient manner and can be interpreted as a human's impulse response to a given behavioral feature for classification.

\section{Opportunities and Challenges at the Intersection of Neurosymbolic Learning and Science}\label{sec:opp}

We have defined and outlined benefits of the NP framework.
However, gaps remain between current NP approaches and practical use cases in science (Figure \ref{fig:process_example}). We draw attention to these challenges to encourage the research community to collaborate in the development of new NP methods to increase the synergy with scientific workflows to accelerate scientific discovery.

\subsection{Dealing with raw, noisy, and imperfect data}\label{sec:data}
Data found in scientific domains provides an opportunity to study NP models with imperfect data in real-world conditions, such as with missing data, experiment noise, and distribution shifts. By incorporating prior knowledge and known constraints in NP models, they have the potential to perform well in the presence of imperfect data. For example, for behavior analysis, neurosymbolic models can automatically learn weak labels from a small amount of annotated examples and apply these trained models to generate weak supervision for a full dataset~\citep{tseng2021automatic}. 

These types of imperfect data exist throughout science: missing data in neural recordings due to hardware issues, noise in pose estimators for tracking animal movements, and distribution shifts. An additional source of noise in data is the considerable variability that exists in the labeling generation process, such as annotator subjectivity and ambiguity in category definitions. 
Furthermore, scientists are often interested in studying rare categories, such as behaviors that may occur in less than 1\% of a dataset.
NP research ~\citep{shah_learning_2020,cui2021differentiable} leverages the flexibility of neural networks with symbolic domain knowledge; however, there remain challenges in improving model scalability that we have outlined in this section. 

\textbf{Structural discovery.} In many scientific workflows, meaningful categories, and structures in raw data may not be clear ahead of time and requires unsupervised or self-supervised learning from data. For example, there are many tools for discovering new behavior categories from data without expert supervision~\citep{pereira2020quantifying}.~\citet{zhan2021unsupervised} demonstrated that integrating domain knowledge in an NP workflow results in more meaningful discovered categories compared to fully neural methods. 
In addition to the algorithmic challenges discussed in previous sections, future research work needs to be robust to variations in experimental noise and produce  interpretations of discovered structures in the data that are useful in the context of science.

\textbf{Distribution Shifts.}   
Distribution shifts are common in real-world applications ~\citep{koh2021wilds}. For typical black-box machine learning models, it is difficult to diagnose and address these errors. NP approaches generally learn interpretable and modular programs, which have the promise to tackle this challenge. For example, in behavior analysis, when the physical behavioral area changes in size, the relative size of mice also changes. This causes errors in behavior classifiers trained in a previously known area, but NP programs can be scaled accordingly to adjust to the new task.

\vspace{-0.1in}
\subsection{Encoding and Learning Domain Knowledge}\label{sec:learning_domain_knowledge}
\vspace{-0.05in}
The success of NP techniques often depends on how a DSL is defined. However, it is not always clear how to handcraft domain-specific components that work best in a scientific context, and this can be a time-consuming process. \emph{Library learning} proposes algorithms that consolidate common patterns in successful programs and add them iteratively to the current DSL, enabling the program search to discover high-performing programs with little effort. 

\textbf{Library learning for science}. 
In behavior analysis, humans are capable of writing short programs that can improve model learning, such as by designing features and heuristics \citep{segalin2021mouse, tseng2021automatic, eyjolfsdottir_detecting_2014}. However, these programs are greatly limited by their simplicity and may not capture complex behavior. Library learning has the potential to augment human feature design, by synthesizing interpretable programs and inducing high-level DSLs, given low-level, generic primitives. 
For example, Dreamcoder~\citep{ellis_dreamcoder_2021} is a library learning system that has been applied to physics equation discovery. 
Library learning has also been studied for generative modeling in molecular chemistry \citep{guo2021data}, which was demonstrated to be able to handle data-limited settings often found in science.

\textbf{Challenges of library learning for science}. In general, it is unclear how library learning can scale to more complex real-world data scientific domains, such as behavior analysis, which often consists of thousands of video frames with noisy data. In addition, it is highly expensive to collect behavior annotations across up to hundreds of behaviors, which is needed to perform traditional library learning. In contrast, current library learning methods have been applied to contexts where each task consists of a few examples, not exceeding hundreds of data points. In addition, the labels are noiseless, as opposed to real-world situations found in behavior analysis \citep{leng_quantifying_2020, segalin2021mouse}. 

Another challenge is that domain experts still need to interpret solutions generated by the NP library learning system. One promising approach leverages natural language to impose a stronger prior on the program search and the library learning \citep{wong2021leveraging}, resulting in a more human-interpretable DSL. Additionally, building a smooth interface between expertise in science, program synthesis, neural networks, and probabilistic library learning methods, found in NP,  would likely require significant engineering and research efforts (Section~\ref{sec:tools}).

\textbf{Representing informal scientific theories.} There is a vast body of knowledge that has been accumulated throughout the span of a given scientific field. Such informal knowledge may not be explicitly represented as a DSL; for instance, behavioral neuroscientists have collected ethograms \citep{garner_mouse_nodate}, or natural language descriptions of the functions of species-specific behavior. Other examples include causal relations between phenomena or interventions in an experimental setting. Past work has proposed logical languages capable of representing intuitive theories of causalities \citep{goodman_learning_2011}. However, capturing all informal and formal knowledge with a single DSL and searching over this space of programs would likely be intractable. Rather, ongoing research in NP focuses on identifying the domain knowledge relevant to a specific subset of scientific problems and distilling such theories into a DSL. 

\subsection{Scalability challenge}

From an optimization standpoint, compared to conventional deep learning, the main additional challenge is searching over program architectures.  Architecture search is in general very challenging and typically leads to combinatorial discrete search space.

\textbf{Inductive synthesis.} A large body of works on program synthesis has focused on \emph{inductive synthesis}, or synthesizing programs from examples \citep{lau_programming_1998, gulwani_automating_2011, devlin_robustfill_2017}. While such a goal is on the surface similar to performing machine learning (ML) with programs as models, a key difference is that ML approaches depend on defining a clear space of models (i.e. neural networks, support vector machines, decision trees) and generalizing to unseen data. In contrast, much work in inductive synthesis considers an arbitrary space of programs and spends significant effort on sample engineering, treating them as noiseless specifications. As a result, inductive synthesis scales poorly with an increase in program length and number of examples. 

\textbf{Scaling NP in science.} 
To tackle scalability in science, models need to handle large and potentially noisy datasets, high-dimensional input space, and a variety of analysis tasks.
Recently, NP research \citep{shah_learning_2020, cui2021differentiable} propose frameworks that scale to large datasets given an expressive DSL. These works are instantiated in behavior analysis: learning programs on temporal trajectory data to reproduce expert annotations of behavior that contain noisy labels, similar to other scientific data. 
These works tackle the challenge of discovering programs with parameters, which can be directly optimized through popular gradient optimization techniques.
While NP methods provide a means of scaling inductive synthesis to scientific datasets, these techniques often involve combining a discrete search over an exponential space of programs with continuous optimization.

\textbf{Challenges for enabling scalability}.  Scaling up program synthesis for neurosymbolic programming is an active field of research.  For instance, differentiable program synthesis methods~\citep{cui2021differentiable} have studied the tradeoff between computation and memory, with heuristics to mitigate memory usage. 
However, training fully neural models on a GPU is often more efficient than training NP models, which requires searching through an exponentially ample space of symbolic architectures on a CPU. Furthermore, scalability has not been broadly explored for different types of scientific data, such as video recordings, which are much higher dimensional than trajectory data.
Finally, the effectiveness of program synthesis may still be limited by the expressivity of a DSL, which requires experts to spend time encoding domain knowledge, such as expert-designed behavior attributes~\citep{sun_task_2021} and temporal filters~\citep{tjandrasuwita2021interpreting} (further discussed in Section~\ref{sec:learning_domain_knowledge}). 

Scalability challenges also arise in other work on symbolic regression  and interpretable machine learning.  For instance, \citet{cranmer_discovering_2020} aims to learn exact mathematical relationships between variables by searching a space of mathematical expressions. 
As another example, \citet{ustun2017optimized} aim to learn optimized risk scores within the same modeling language used by clinicians, which leads to an NP-hard optimization problem that they solve using integer programming techniques.

\subsection{Challenges of optimization of discrete and continuous space in neurosymbolic programs}\label{sec:optimization}
NP relies on techniques from symbolic program synthesis to facilitate interpretable and verifiable searches over the scientific hypothesis space. However, programs are inherently symbolic, owing to their roots in mathematical logic. This makes modeling phenomena in the continuous domain challenging without modifying the way we interpret programs. 

For instance, consider a simple program that is modeled by an if-then-else statement (\texttt{if condition do expr1 else do expr2}). The possible behaviors of \texttt{condition} are partitioned into two sets -- True (\texttt{1}) or False (\texttt{0}). These sets evaluate to either \texttt{expr1} or \texttt{expr2} respectively. However, an NP approach requires reasoning to be differentiable over a \textit{gradient} of possibilities. Discrete programs are inaccurate models for these applications. Specifically, in behavior classification, modeling the ``attack'' action using a symbolic if-then-else expression would partition the mouse's aggression into a binary set: either always attacking or not attacking at all. What makes more sense is to model ``attacking'' as a binomial distribution. This requires \textit{relaxing} our symbolic if-then-else to account for a continuous gradient of probabilities from 0\% to 100\%.

\textbf{Continuous relaxations.} We approach the continuous program optimization problem of the symbolic domain by changing the semantics of the programming language. Specifically, work on \textit{Smooth Interpretation} \citep{ChaudhuriS10} rewrites discrete functions using their closest smooth mathematical functions. Consecutively, an if-then-else statement would be rewritten as a hyperbolic tangent function with a high temperature. This smoothening is not restricted to a one-dimensional input space and specialized functions. In general, in higher dimensions, we can use Gaussian smoothing to smooth discontinuities. 
Such relaxations, in conjunction with other program analysis tools, allow gradient descent-based optimizers to converge to optimal programmatic models.

Continuous relaxations enable an approximate interface between neural networks and programming languages, which are essential in the NP framework. For example, in Houdini \citep{valkov2018houdini}, continuous relaxations enabled the construction of a functional programming language that admits neural networks and higher-order functions. This construction facilitated the high-level transfer of learned concepts across tasks in a lifelong learning setting. In NEAR \citep{shah_learning_2020}, the interface between neural networks and differentiable programs allowed for measuring the performance of partial programs. This proved to be an $\epsilon-$admissible heuristic for synthesizing differentiable programs in the behavior analysis setting.

Smooth Interpretation allows positing a differentiable \textit{approximation} for a non-differentiable program. This approximation error introduces a tradeoff between the output precision and optimal trainability of the model. Specifically, under-approximating the non-differentiable components might increase the precision of the differentiable program at the cost of retaining discontinuities in the optimization landscape and converging to a suboptimal model, and vice-versa. 

\subsection{Evaluating Interpretability}\label{sec:evaluation}
The main goal of interpretability is to obtain insights that are understandable and actionable to humans and to assist scientists in their analysis workflow. The following are commonly described properties of explanations found in machine learning \citep{pml2Book}, that have the potential to improve the interpretability and evaluation of NP workflows: \textbf{Compactness or sparsity:} Sparsity generally corresponds to some notion of smallness measurement (a few features or a few parameters); \textbf{Completeness:} To measure if the explanation includes all the relevant elements,  higher-level
concepts needed; \textbf{Stability:} To measure the extent that there are explanations similar for similar input; \textbf{Actionability:} To allow focusing on only aspects of the model that the user might be able to intervene on; \textbf{Modularity:} Explanation can be broken down into understandable parts.
To study interpretability of NP models for science, we need datasets and benchmarks to quantify these different dimensions of interpretability across scientific contexts, which is currently an open problem.

An example of an interpretable program is presented in \autoref{fig:Np2}. This neurosymbolic program classifies the ``sniff" action between two mice. An interpretation is that if the distance between two mice is small, they are doing a ``sniff"; otherwise, they are only doing a ``sniff" if the accelerations are small.
Interpretability can also be expressed as a utility function or a form of abstraction \cite{stitch2023}
which minimizes the size of the corpus of a neurosymbolic program.
\subsection{Cross-Domain Benchmarking}\label{sec:benchmarking}
While many individual fields of science have seen some successes through NP, consolidating underlying generalizable and cross-cutting insights remains another significant open challenge for the scientific and machine learning communities. 
Towards this, we propose to build initial benchmarks around low-dimensional spatiotemporal data, a setting where NP methods have demonstrated potential~\citep{shah_learning_2020,verma2018programmatically}. 
We believe that there are several benefits to gain from developing an NP benchmark for the ML and scientific communities: (1) systematic improvements across broad scientific use cases, (2) comprehensive model evaluations, instead of in domain-specific dimensions, (3) increased awareness of important scientific applications that have not received as much attention from the ML community.

\textbf{Challenges of benchmarking NP for science.} Interpreting programmatic structures requires expert domain knowledge, which can be expensive and time-consuming to obtain. In behavior analysis, evaluating learned programmatic structures requires interactions with experts in the behavioral science community. This imposes a major bottleneck on evaluating outputs. 
A standardized benchmark will make it easier for the community to convene and interact with a panel of experts. We believe that developing a benchmark for NP pipelines is integral to moving the NP field forward.

The space of NP models is broad. Each algorithm presents a unique methodology for encoding expert knowledge into the NP lifecycle. This requires comparing models on multiple evaluation metrics. However, not all NP algorithms can be systematically evaluated on the same set of metrics. For instance, certain classes of models use stochastic search to discover the programmatic structure and the programs found by such an approach may not be reproducible. Additionally, NP models might exhibit properties that do not have concrete evaluation metrics. 
For instance, classes of NP algorithms that exhibit robust reproducibility. That is, the model's outputs are reproducible with small perturbations to the input data. 
However, to the best of our knowledge, defining such a metric quantitatively and objectively remains an open challenge.

The hardware requirements for learning neural representations and symbolic functions are orthogonal. Neural network training is GPU intensive, while program synthesis is CPU intensive. This increases the cost of computation and imposes a barrier to entry for aspiring NP researchers. NP benchmarks need to take the efficiency and performance of training and inference into account.

\subsection{Cross-Domain Analysis Tools for Scientists}\label{sec:tools}

\begin{figure}
    \makebox[\textwidth][c]{\includegraphics[scale=0.35]{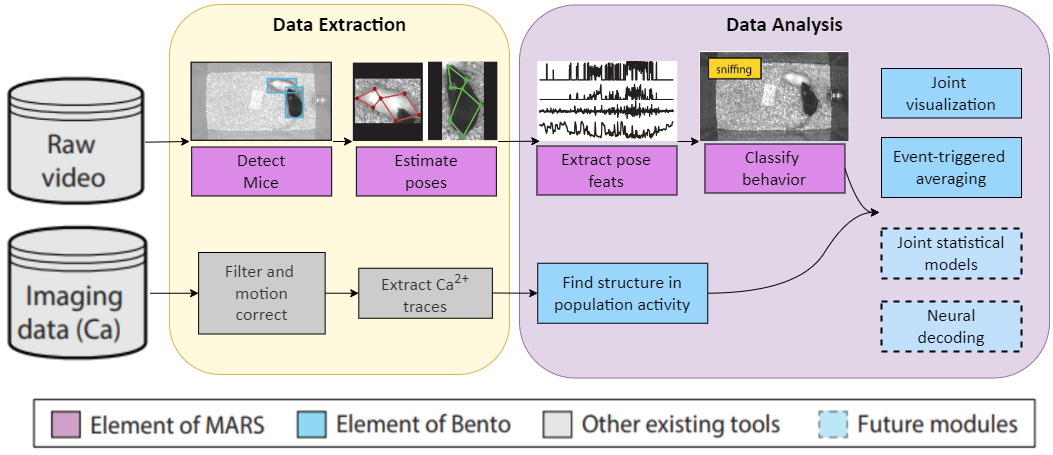}}
    \caption{Functionalities of MARS and Bento \citep{segalin2021mouse} in the behavior analysis pipeline.}
    \vspace{-0.2in}
    \label{fig:mars_pipeline}
\end{figure}

\textbf{Importance of tools in science.} User-friendly tools are important for facilitating the integration of ML models in real-world science workflows but have not been well-explored for NP approaches. 
For example, numerous tools, based on statistical analysis and ML, have been developed to interface with scientists and facilitate behavior analysis from videos in~\citet{pereira2020quantifying}.
These tools assist with much of the computational pipeline for behavior classification as outlined in Figure \ref{fig:mars_pipeline}, and will often provide visual interfaces that visualize relevant raw data such as video, with model outputs, such as pose data and behavior~\citep{segalin2021mouse}.
Enabling similar tools for NP approaches has the potential to benefit existing scientific workflows. For instance, integrating NEAR into a visual interface could provide scientists with a user-friendly way of generating differentiable programs and means of understanding the programs from NP pipelines. The parameters associated with programmatic primitives are likely to have a much more human interpretation \citep{tjandrasuwita2021interpreting} than those found in black-box neural networks.

\textbf{Challenges of building NP tools.}  Domain expertise in science varies in structure, from behavioral attributes to visual or textual descriptions, to known dynamics of movement, to knowledge graphs, and generally differs across labs and domains. 
Furthermore, to measure progress, user evaluations are needed that could offer  quantitative or qualitative evidence in NP workflows. 
Taking the first steps to realize and evaluate the effectiveness of NP algorithms through a human-computer interaction approach may not only improve the scientific pipeline but also yield new algorithmic directions on combining NP with more traditional human-in-the-loop methods, such as active learning.

\section{Conclusion}
Neurosymbolic programming offers the promise to accelerate scientific discovery and optimize scientific discovery end-to-end. The benefits are in its ability to incorporate prior knowledge and the symbolic nature of the solutions, essential scientific workflows.  However, challenges still remain in scalability and optimization stability of these approaches, comprehensive evaluations, and deployment in the form of tools. In this paper, we have demonstrated the opportunities and challenges of neurosymbolic programming in a concrete scientific application, behavior analysis.
A key promise of neurosymbolic programming is to provide a set of unifying principles in interpretable machine learning and prior scientific literature.
We invite the science and computer science communities to adopt these methods in their scientific workflow and to contribute to the research to advance NP techniques for science due to the unique benefit to these communities.

\textbf{Funding.}
This project was supported by the National Science Foundation under Grant \#1918839 ``Understanding the World Through Code'' http://www.neurosymbolic.org/

\bibliographystyle{plainnat}
\bibliography{reference}

\end{document}